\documentclass[letterpaper, 10 pt, journal,twoside]{URL-ieeeconf}
\IEEEoverridecommandlockouts    

\usepackage{graphics}           
\usepackage{times}              
\usepackage{amsmath}            
\usepackage{amssymb}            
\usepackage{graphicx}
\usepackage{algorithm}
\usepackage[noend]{algpseudocode}
\usepackage{booktabs}
\usepackage{color}
\usepackage{multirow}
\usepackage{subcaption}
\usepackage{rotating}
\usepackage{cite}
\usepackage{hyperref}
\usepackage{soul}
\usepackage{tikz}
\usetikzlibrary{calc}
\hypersetup{
	colorlinks=true,
	linkcolor=black,
	filecolor=black,      
	urlcolor=black,
	pdftitle={Overleaf Example},
	pdfpagemode=FullScreen,
}
\definecolor{instructioncolor}{rgb}{.5,.5,.5}

\def\secref#1{Section~\ref{#1}}
\def\figref#1{Fig.~\ref{#1}}
\def\tabref#1{Table~\ref{#1}}
\def\eqref#1{(\ref{#1})}

\def\vsfig{\vspace{-0.4cm}}
\def\vstab{\vspace{-0.4cm}}
\def\vsequ{\vspace{-0.15cm}}

\newcommand{\rom}[1]{\uppercase\expandafter{\romannumeral #1\relax}}

\makeatletter
\usepackage{xspace}
\DeclareRobustCommand\onedot{\futurelet\@let@token\@onedot}
\def\@onedot{\ifx\@let@token.\else.\null\fi\xspace}

\def\etal{{\textit{et al}}\onedot}
\makeatother

\usepackage{array}
\newcolumntype{L}[1]{>{\raggedright\let\newline\\\arraybackslash\hspace{0pt}}m{#1}}
\newcolumntype{C}[1]{>{\centering\let\newline\\\arraybackslash\hspace{0pt}}m{#1}}
\newcolumntype{R}[1]{>{\raggedleft\let\newline\\\arraybackslash\hspace{0pt}}m{#1}}

\definecolor{rvc}{RGB}{0, 0, 255}
\definecolor{cfv2}{RGB}{0, 0, 0}

\makeatletter
\newif\if@anonymize

\@anonymizetrue    

\if@anonymize
\newcommand{\highlight@DoHighlight}{
	\fill [outer sep = -15pt, inner sep = 0pt, color=black]
	($(begin highlight)+(0,8pt)$) rectangle ($(end highlight)+(0,-3pt)$) ;
}

\newcommand{\highlight@BeginHighlight}{
	\coordinate (begin highlight) at (0,0) ;
}

\newcommand{\highlight@EndHighlight}{
	\coordinate (end highlight) at (0,0) ;
}

\newdimen\highlight@previous
\newdimen\highlight@current
\newlength{\item@width}

\DeclareRobustCommand*\anonymize{%
	\SOUL@setup
	\def\SOUL@preamble{%
		\begin{tikzpicture}[overlay, remember picture]
		\highlight@BeginHighlight
		\highlight@EndHighlight
		\end{tikzpicture}%
	}%
	\def\SOUL@postamble{%
		\begin{tikzpicture}[overlay, remember picture]
		\highlight@EndHighlight
		\highlight@DoHighlight
		\end{tikzpicture}%
	}%
	\def\SOUL@everyhyphen{%
		\discretionary{%
			\SOUL@setkern\SOUL@hyphkern
			\SOUL@sethyphenchar
			\tikz[overlay, remember picture] \highlight@EndHighlight ;%
		}{%
		}{%
			\SOUL@setkern\SOUL@charkern
		}%
	}%
	\def\SOUL@everyexhyphen##1{%
		\SOUL@setkern\SOUL@hyphkern
		\settowidth{\item@width}{##1}%
		\makebox[\item@width]{}%
		\discretionary{%
			\tikz[overlay, remember picture] \highlight@EndHighlight ;%
		}{%
		}{%
			\SOUL@setkern\SOUL@charkern
		}%
	}%
	\def\SOUL@everysyllable{%
		\begin{tikzpicture}[overlay, remember picture]
		\path let \p0 = (begin highlight), \p1 = (0,0) in \pgfextra
		\global\highlight@previous=\y0
		\global\highlight@current =\y1
		\endpgfextra (0,0) ;
		\ifdim\highlight@current < \highlight@previous
		\highlight@DoHighlight
		\highlight@BeginHighlight
		\fi
		\end{tikzpicture}%
		\settowidth{\item@width}{\the\SOUL@syllable}%
		\makebox[\item@width]{}%
		\tikz[overlay, remember picture] \highlight@EndHighlight ;%
	}%
	\SOUL@
}
\else
\newcommand{\anonymize}[1]{#1}
\fi
\makeatother

\newcommand{\RR}{\mathbb{R}}

\newcommand{\normal}{\mathbf{n}}

\renewcommand{\b}[1]{\mbox{\boldmath$#1$}}

\newcommand{\inv}{^{-1}}

\newcommand{\mrb}{\b {\mathrm{b}}}

\newcommand{\mrn}{\b {\mathrm{n}}}

\newcommand{\mrp}{\b {\mathrm{p}}}

\newcommand{\mrr}{\b {\mathrm{r}}}
\newcommand{\mrs}{\b {\mathrm{s}}}

\newcommand{\mrv}{\b {\mathrm{v}}}

\newcommand{\mrx}{\b {\mathrm{x}}}

\newcommand{\mrH}{\b {\mathrm{H}}}
\newcommand{\mrI}{\b {\mathrm{I}}}
\newcommand{\mrJ}{\b {\mathrm{J}}}

\newcommand{\mrR}{\b {\mathrm{R}}}

\title{\LARGE \bf LVI-Q: Robust LiDAR-Visual-Inertial-Kinematic Odometry for Quadruped Robots Using Tightly-Coupled and Efficient Alternating Optimization}

\author{Kevin Christiansen Marsim$^{1}$, Minho Oh$^{1}$, Byeongho Yu$^{2}$,  Seungjae Lee$^{1}$, I Made Aswin Nahrendra$^{1}$, \\ Hyungtae Lim$^{1}$, \textit{Member, IEEE}, and Hyun Myung$^{*1}$, \textit{Senior Member, IEEE}
\thanks{Manuscript received: May 21, 2025; Revised June 4, 2025; Accepted July 22, 2025.}
\thanks{This paper was recommended for publication by Editor Pascal Vasseur upon evaluation of the Associate Editor and Reviewers’ comments. This work was supported in part by the Korea Evaluation Institute of Industrial Technology (KEIT)~(20018216, Development of mobile intelligence SW for autonomous navigation of legged robots in dynamic and atypical environments for real-world applications) and in part by the National Research Foundation of Korea~(NRF)~(RS-2024-00348461) funded by the Korean Government (MOTIE \& MSIT). The students are supported by the BK21 FOUR~(Republic of Korea).}
\thanks{$^*$Corresponding author: Hyun Myung}
\thanks{$^{1}$The authors are with the School of Electrical Engineering, KAIST (Korea Advanced Institute of Science and Technology), Daejeon, 34141, Republic of Korea. {\tt\scriptsize \{kevinmarsim, minho.oh, sjlee, anahrendra, shapelim, hmyung\}@kaist.ac.kr}}
\thanks{$^{2}$The author is with URobotics Corp., Daejeon, 34051, Republic of Korea. {\tt\scriptsize bhyu@urobotics.ai}
\hfill}
}
\begin{document}
\maketitle
\begin{abstract}
  %
  Autonomous navigation for legged robots in complex and dynamic environments relies on robust simultaneous localization and mapping~(SLAM) systems to accurately map surroundings and localize the robot, ensuring safe and efficient operation. While prior sensor fusion-based SLAM approaches have integrated various sensor modalities to improve their robustness, these algorithms are still susceptible to estimation drift in challenging environments due to their reliance on unsuitable fusion strategies. 
  Therefore, we propose a robust \mbox{LiDAR-visual-inertial-kinematic} odometry system that integrates information from multiple sensors, such as a camera, LiDAR, inertial measurement unit (IMU), and joint encoders, for visual and LiDAR-based odometry estimation.
  Our system employs a fusion-based pose estimation approach that runs optimization-based visual-inertial-kinematic odometry (VIKO) and filter-based LiDAR-inertial-kinematic odometry (LIKO) based on measurement availability. In VIKO, we utilize the foot-preintegration technique and robust LiDAR-visual depth consistency using superpixel clusters in a sliding window optimization. In LIKO, we incorporate foot kinematics and employ a point-to-plane residual in an error-state iterative Kalman filter~(ESIKF). Compared with other sensor fusion-based SLAM algorithms, our approach shows robust performance across public and long-term datasets.
  
\end{abstract}

\vspace{-3mm}
\section{Introduction}
\label{sec:intro}

\PARstart{A}{dvancements} in robotics have motivated the development of legged robots renowned for their agility and compact design. Legged robots are equipped with joint encoders to facilitate the estimation of their current stance. A key research focus for these robots is autonomy for tasks such as the exploration of unknown environments~\cite{miller2020ral} and the execution of intricate manipulations~\cite{hooks2020ral}.

Robust localization or state estimation of a robot, specifically legged robots in our case, is crucial for achieving a high level of autonomy~\cite{wisth2023tro}.
Researchers have explored sensor fusion-based approaches to enhance the robustness of localization and pose estimation by leveraging different sensor modalities.
Consequently, various methods have been proposed for \mbox{visual-inertial} odometry~(VIO), \mbox{LiDAR-inertial} odometry~(LIO), and \mbox{LiDAR-visual-inertial}~(LVI) odometry~\cite{shan2021icra}.

\begin{figure}[t!]
	\captionsetup{font=footnotesize}
	\centering
	\def\svgwidth{0.43\textwidth}
	\graphicspath{{pics/0919_Motivation/}} 
	\input{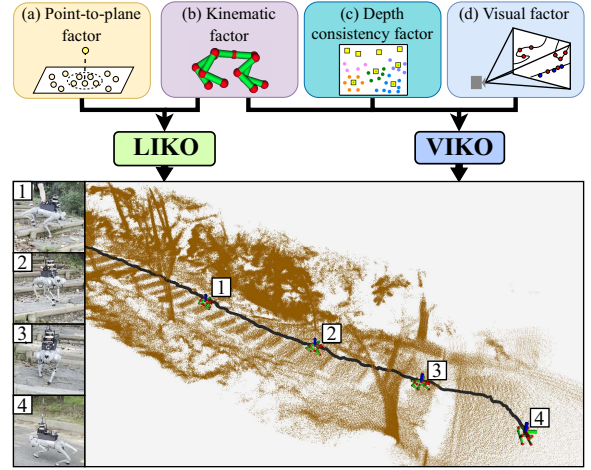}
	\caption{Overview of our odometry system that combines multiple modalities from different sensors in the form of (a) point-to-plane factor where we minimize the distance from the current LiDAR scan to the map plane, (b) kinematic factor obtained from foot-preintegration method, (c) depth consistency factor by combining visual features and LiDAR point cloud, and (d) visual factor from tracking unique features on the image.}
	\label{fig:motivation}
	\vsfig
\end{figure}
However, each of these methods has specific limitations during deployment. First, VIO, which optimizes states based on visual feature reprojection errors or direct photometric errors between images~\cite{qin2018tro,song2022ral}, is prone to failure when there are significant changes in illumination or aggressive motions. Second, LIO, which minimizes the error of each measured point in a previously or simultaneously constructed point-cloud map~\cite{wei2022tro,shan2020iros}, often experiences degeneracy in odometry when the surrounding areas have similar geometric features, such as long corridors~\cite{lim2023ur}.

\mbox{Factor-graph} and \mbox{filter-based} LVI approaches have been proposed to address the limitations of VIO and LIO~\mbox{\cite{lin2022icra, wisth2023tro, zheng2022iros}}. These methods handle LiDAR and visual measurements in a tightly-coupled manner to enhance accuracy and reduce degeneracy caused by measurement failures. 
However, these LVI fusion strategies still struggle in extreme conditions, such as rapid motion and inconsistent illumination, that cause misalignment between visual and \mbox{LiDAR} measurements~\cite{zhao2021iros,khattak2020icuas}. 





We tackle this issue by developing a novel odometry system that incorporates additional modality for both VIO and LIO processes in a unified estimation framework. In particular, kinematic data from joint sensors is integrated into VIO and LIO to constrain the robot motion estimation. The additional information can be used as an extra constraint to prevent the result from diverging due to lack of features or unstable information.

In this letter, we propose~\textit{LVI-Q}, which is an abbreviation of~\textit{LiDAR-Visual-Inertial odometry for Quadrupeds}, that addresses the limitations of existing LVI approaches by tightly coupling \mbox{LiDAR-visual-inertial-kinematic} data into the alternating estimation process that combines the strengths of factor-graph and filter-based optimization processes, as shown in~\figref{fig:motivation}. First, we propose a LiDAR-inertial-kinematic
odometry~(LIKO) module consisting of error-state iterative Kalman filter (ESIKF) that utilizes the \mbox{point-to-plane}~(\figref{fig:motivation}(a)) residual and kinematic factor~(\figref{fig:motivation}(b)). 
We chose the filter-based approach for LIKO module due to its low latency, typically under 20~ms~\cite{xu2021ral}, compared with that of factor-graph-based LIO systems, typically over 50~ms and may increase further as the map size grows~\cite{shan2020iros}. Additionally, the ESIKF linearizes the error state, which behaves more linearly than the full (nominal + error) state.
In addition to our previous work~\cite{yeeun2022icra}, we further develop foot-preintegration residual to improve the filter-based method. 
Second, the visual-inertial-kinematic odometry~(VIKO) module is proposed to improve the trajectory using sliding window optimization with kinematics~(\figref{fig:motivation}(b)), depth consistency~(\figref{fig:motivation}(c)), and visual factors~(\figref{fig:motivation}(d))~\cite{qin2018tro}. 
We chose the factor-graph approach in VIKO to further optimize past states within the sliding window, including visual feature depth through the depth consistency factor, thereby ensuring that the estimation of the current states are consistent. 


\begin{figure*}[t!]
	\vspace{0.4cm}
	\captionsetup{font=footnotesize}
	\centering
	\def\svgwidth{0.90\textwidth}
	\graphicspath{{pics/0410_Framework/}} 
	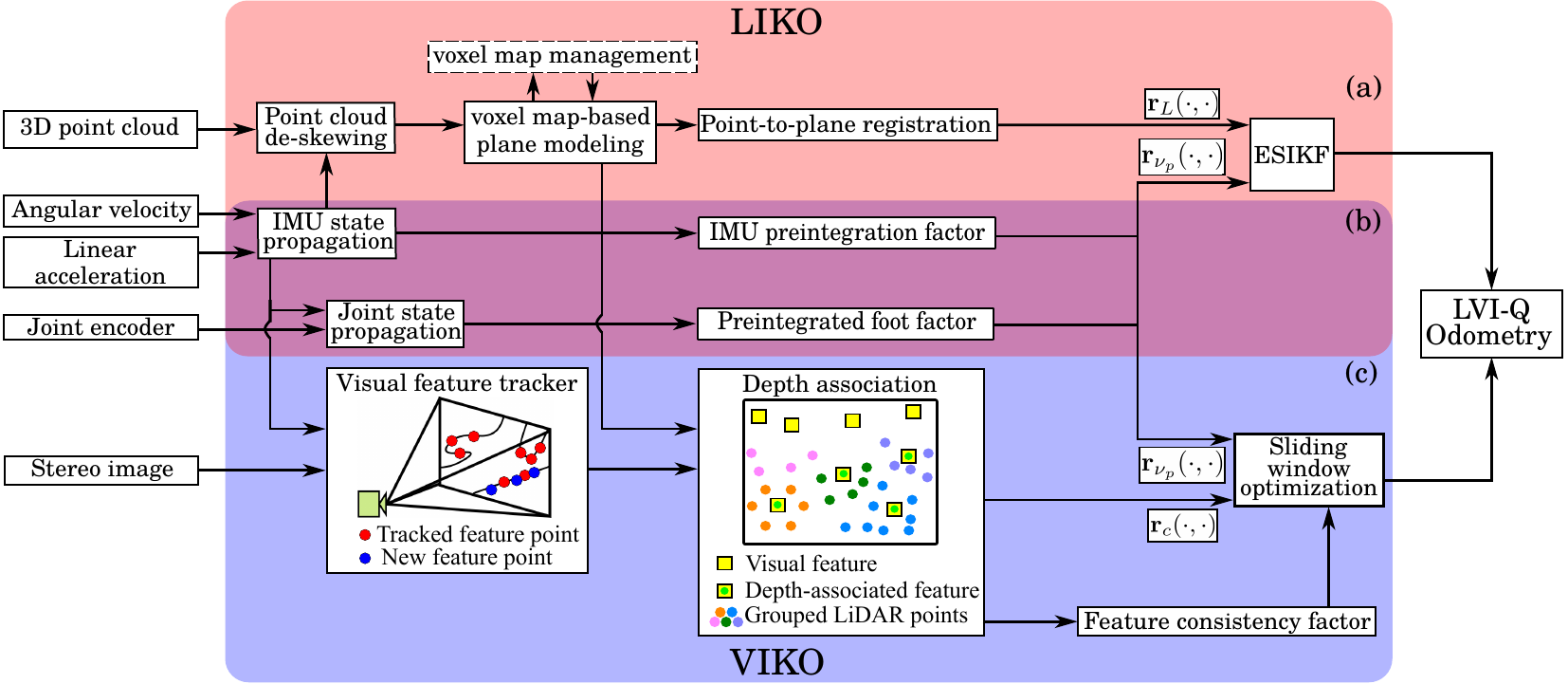
	\caption{The overall framework of LVI-Q with LIKO and VIKO which construct three major factors optimized in an alternating estimation framework comprising of error state iterative Kalman filter~(ESIKF) and sliding window optimization. (a) Construction of the LiDAR point-to-plane residual through efficient voxel map management and plane modeling. (b) The kinematic factor determines the robot's stance by leveraging joint state propagation. Preintegrated foot measurements are utilized to optimize the current body pose. (c) Integration of the visual factor by tracking prominent image features and the depth consistency module derived from LiDAR point cloud data.}
	\label{fig:system_and_FactorGraph}
	\vspace{-0.4cm}
\end{figure*}

The main contributions of this letter are threefold:
\begin{enumerate}
	\item We propose a VIKO module that utilizes kinematic and visual feature depth consistency factors. The depth consistency factor is based on \mbox{superpixel-grouped} 3D point cloud distribution fitting to ensure consistent scale tracking during sliding window-based odometry estimation.
	\item We propose a LIKO module that couples foot-preintegration residuals with LiDAR information in a tightly-coupled manner, using filter-based estimation, as an additional constraint in challenging scenes.
	\item We demonstrate the capabilities of the LVI-Q's alternating optimization process by testing it on various datasets across multiple platforms.
\end{enumerate}


%
%
%
%


\section{Related Work}
\label{sec:related}

\vspace{-1mm}

\subsection{Inertial-Kinematic State Estimation}
\label{sec:related_legged}

Bloesch \textit{et al.}~\cite{bloesch2013robotics} proposed an odometry algorithm that fused inertial measurement unit (IMU) and kinematic data using an Unscented Kalman Filter. Camurri \textit{et al.}~\cite{camurri2020frontier} combined IMU data with velocity estimates from foot kinematics. They also proposed a contact detector that adaptively adjusts the measurement covariance to compensate for unreliable kinematic factors. Yang \textit{et al.}~\cite{yang2022icra} proposed a visual-inertial-kinematic odometry algorithm that simultaneously performs online kinematic parameter calibration to reduce position drift. 

These studies assume static foot contact and model foot slip as Gaussian noise (non-slip assumption)~\cite{ross2018ral}, which is often violated in unstructured environments with frequent slippage. As a result, such methods require precise contact detection. In contrast, LVI-Q bypasses the non-slip assumption, enabling robust estimation in unstructured environments without requiring a contact detector.

\subsection{LiDAR-Visual-Inertial State Estimation}
\label{sec:related_lvi}
Many studies integrate information from sensors, such as cameras and LiDARs, with data from inertial sensors because they complement each other effectively.
The authors in~\cite{brossard2018iros} utilized the Kalman filter scheme to combine IMU data with vision data. Similarly, Fast-LIO2~\cite{wei2022tro} and \mbox{Faster-LIO}~\cite{bai2022ral} exploit LiDAR point cloud based on a Kalman filter scheme.
Unlike these algorithms, some methods optimize visual feature points~\cite{qin2018tro} or LiDAR point clouds~\cite{shan2020iros} using a sliding window scheme to reduce computational cost.

Incorporating LiDARs and cameras has been widely considered to achieve more accurate results.
LVI-SAM~\cite{shan2021icra} extends LIO-SAM~\cite{shan2020iros} by fusing visual data through factor graph-based smoothing and mapping. R2LIVE~\cite{lin2021ral} and R3LIVE~\cite{lin2022icra} employ a tightly-coupled approach, merging LiDAR, visual, and inertial data, using an ESIKF framework.
However, these methods rely on a single fusion strategy that processes all modalities simultaneously, limiting their adaptability to challenging conditions where one sensor type may dominate or fail.

In contrast, our LVI-Q system introduces an alternating optimization framework that dynamically switches between filter-based and factor-graph-based optimizations, leveraging the strength of each approach based on real-time sensor availability to enhance robustness and accuracy. Additionally, we propose a novel method to incorporate LiDAR to camera measurements by aligning visual feature points with surrounding point cloud distribution. This method provides benefits, particularly in cases where the camera may yield poor measurements, by ensuring the reliability of 3D visual feature points.

\subsection{LiDAR-Visual-Inertial-Kinematic State Estimation} 
Utilizing multiple exteroceptive and proprioceptive sensors enhances the system's robustness against failures or partial degradations in individual sensors. Inspired by this, Wisth~\textit{et al.}~\cite{wisth2023tro} proposed VILENS, which jointly optimizes visual, LiDAR, inertial, and kinematic information within a single factor graph. Additionally, they introduced a linear velocity bias to address
slippage, deformations, and other kinematic inaccuracies at
the contact point.

Similarly, we also use all available sensor information from multiple modalities. However, we directly exploit kinematic information via~\mbox{foot-preintegration}~\cite{yeeun2022icra}, which allows the kinematic factor to be utilized in every optimization step rather than only during stable foot contact. In the case of visual or LiDAR measurement degeneracies, LVI-Q can compensate for the error by relying on the kinematics factor. 


\section{Proposed Method}
\label{sec:main}
This section explains the proposed LVI-Q system which estimates the robot motion according to the measurements from the IMU, joint, camera, and LiDAR sensors with a unified estimation approach. \figref{fig:system_and_FactorGraph} provides an overview of our proposed method, comprising of three primary factors: a) LiDAR, b) foot-preintegration, and c) visual factors. The VIKO and LIKO modules construct visual-kinematic and LiDAR-kinematic factors, respectively. We alternate between VIKO and LIKO modules based on the available visual or point cloud measurements to fully leverage the strengths of both optimization frameworks. The constructed visual-kinematic or LiDAR-kinematic factors are then forwarded to the sliding window optimization or ESIKF block.

\subsection{State Definition}
\label{subsec:state_notation}
For clarity, we introduce the notations used in this letter. Reference frames are denoted as $(\cdot)^b$, $(\cdot)^L$, $(\cdot)^{c_r}$, $(\cdot)^{c_l}$, and $(\cdot)^w$ for the body centered at the IMU, LiDAR, right camera, left camera, and world frames, respectively. The state vector ${\mathrm{\bf{x}}}_i$, when the $i$-th keyframe is received, is represented as follows:
\vsequ
\begin{equation}
\label{states}
{\mathrm{\bf{x}}}_{i} \ = \ [{\mrp}^{w}_{i} \ {\mrR}^{w}_{i} {\ ^{c_l}\mrp}^{b}_{i} \ {^{c_l}\mrR}^{b}_{i} \ {^{c_r}\mrp}^{b}_{i} \ {^{c_r}\mrR}^{b}_{i} \ {\mrv}^{w}_{i} \ {\mrb}^{a}_{i} \ {\mrb}^{g}_{i} ]^{\mathsf{T}},
\end{equation}
where {{\mrp}$^{w}_{i}$$\in{\RR}^3$} and {\mrR}$^{w}_{i}$$\in SO(3)$ represent the current robot position and orientation in the world frame. ${^{c_l}\mrp}^{b}_{i}$$\in{\RR}^3$, ${^{c_l}\mrR}^{b}_{i}$$\in SO(3)$, ${^{c_r}\mrp}^{b}_{i}$$\in{\RR}^3$, and ${^{c_r}\mrR}^{b}_{i}$$\in SO(3)$ are the left and right camera extrinsic parameters with respect to the IMU sensor, respectively. ${\mrv}^{w}_{i}$$\in{\RR}^3$ is the current robot velocity in the world frame. {\mrb}$^{a}_{i}$$\in{\RR}^3$ and {\mrb}$^{g}_{i}$$\in{\RR}^3$ are the accelerometer and gyroscope biases, respectively.

We also estimate the orientation and position of the $\l$-th foot in the world frame, denoted by ${\b \Psi}^{w}_{{l},{i}}$ and ${\mrs}^{w}_{{l},{i}}$, respectively. In the VIKO module, the states are stored in a sliding window~${\b{\mathcal{X}}}$. Additionally, we parameterize the inverse feature depth
, denoted as ${{\lambda}_k}$, where $k$ represents the tracked feature index. This results in an extended VIKO state $^{v}{\mathrm{\bf{x}}}_{i}$ as follows:
\vsequ
\begin{equation}
\begin{aligned}
{\b{\mathcal{X}}} & = \ [^{v}{\mathrm{\bf{x}}}_{0} \ ^{v}{\mathrm{\bf{x}}}_{1} \ \dots \ ^{v}{\mathrm{\bf{x}}}_{i-1} \ {\lambda_0} \ {\lambda_1} \ \dots \ { \lambda_k}], \\
^{v}{\mathrm{\bf{x}}}_{i} &= \ [ {\mathrm{\bf{x}}_i}^{\mathsf{T}} \ {{\b \Psi}^{w}_{{l},{i}}}^{\mathsf{T}} \ {{\mrs}^{w}_{{l},{i}}}^{\mathsf{T}} ]^{\mathsf{T}}.
\end{aligned}
\vspace{-0.5mm}
\end{equation}
The estimated states will be denoted with~$\hat{\left(\cdot\right)}$. We also follow the boxplus and boxminus operations discussed in~\cite{wei2022tro}. Therefore, we can write the estimated states in terms of the true state $\mathbf{x}_i$ and estimation error~${ \delta}\hat{\mathrm{\bf{x}}}_{i}$ as follows:
\vspace{-1mm}
\begin{equation}
\label{eq:estimation_error}
\hat{\mathrm{\bf{x}}}_{i} \ = {\mathrm{\bf{x}}}_{i} \ \boxplus \ {\delta}\hat{\mathrm{\bf{x}}}_{i}.
\vspace{-1mm}
\end{equation}

\subsection{IMU State Propagation}
\label{subsec:imu_prop}
The IMU state propagation scheme initiates a prior motion distribution before optimization, indicating the uncertainty associated with the current state estimation error. The states and covariances are propagated for each IMU measurement according to the specified propagation model defined in~(5) and (10) in~\cite{wei2022tro}.

\subsection{Cost Function}
\label{subsec:cost_function}
Following the IMU state propagation, we optimize the cost functions from the available visual-kinematic or LiDAR-kinematic measurements. 

In the LIKO module, we aim to obtain the \mbox{maximum-a-posteriori}~(MAP) estimate~$\check{\mathrm{\bf{x}}}_{k+1}$ of the current robot state in the ESIKF by minimizing the cost function related to its estimation error $\delta\check{\mathrm{\bf{x}}}_{k+1}$ at the $(k+1)$-th measurement time as follows:
\vspace{-0.12cm}
\begin{equation}
\begin{aligned}
\label{eq:ESIKF_cost_function}
& \min_{\delta\check{\mathrm{\bf{x}}}_{k+1}}\left. \Bigg\{ \left\Vert \check{\mathrm{\bf{x}}}_{k+1} \boxminus \hat{\mathrm{\bf{x}}}_{k+1}+\b{\mathcal{H}}{\delta\check{\mathrm{\bf{x}}}_{k+1}} \right\Vert_{\b \Omega_{{\delta \hat{\mathrm{\bf{x}}}}_{k+1}}} \right. \\ & \left. +\sum_{n=1}^{N_{\mathcal{P}}}\left\Vert \mrr_{\mathcal{P}}\left( \check{\mathrm{\bf{x}}}_{k+1},\b {\mathcal{P}}^w_{n,t_{k+1}} \right) + \mrH^\mathcal{P}_n\delta\check{\mathrm{\bf{x}}}_{k+1} \right\Vert^2_{\b \Omega_{\mathcal{P}}} \right. \\ & \left. + \sum^{N_l}_{l=1}\left\Vert \mrr_{\nu_p}\left(\check{\mathrm{\bf{x}}}_{k+1},\b \nu^l\right)+\mrH^{\nu_p}_l\delta\check{\mathrm{\bf{x}}}_{k+1} \right\Vert^2_{\b \Omega_{\nu_p}}\right\},
\end{aligned}
\end{equation}
where~$\mrr_{\mathcal{P}}(\cdot,\cdot)$ and~$\mrr_{\nu_p}(\cdot,\cdot)$ denote the LiDAR and preintegrated foot velocity residuals, respectively.~$\b \Omega_{\mathcal{P}}$ and~$\b \Omega_{\nu_p}$ are the covariances for each measurement.~$\b{\mathcal{H}}$,~$\mrH^\mathcal{P}_n$, and~$\mrH^{\nu_p}_l$ are the Jacobian matrices of the prior, LiDAR, and preintegrated foot velocity residuals with respect to the estimation error, respectively. $\b {\mathcal{P}}^w_{n,t_{k+1}}$ is the $n$-th \mbox{LiDAR} point deskewed to the current time, as explained in Section~\ref{subsec:liko}. $N_\mathcal{P}$ and $N_l$ denote the numbers of LiDAR points and foot measurements, respectively. Finally, ~$\b{\nu}^l$ is the foot velocity measurement of the $l$-th foot obtained from foot velocity preintegration technique~\cite{yeeun2022icra}. The exact definitions and derivations of the Jacobians can be found in~\cite{lin2021ral}. 

We optimize the states in VIKO using a sliding window optimization with a total number of $\mathbb{I}$ windows, which can be modeled as the MAP estimate $\check{\b{\mathcal{X}}}$ as follows:
\vspace{-1mm}
\begin{equation}
\begin{aligned}
\label{eq:sliding_window_cost_function}
&{\min_{\check{\b{\mathcal{X}}}}} \left\{ \sum_{i\in\mathbb{I}}\sum_{m=1}^{N_c} \left(\rho\left(\left\Vert\mrr_c(\check{\b{\mathcal{X}}},\b{\mathrm{f}}_m)\right\Vert^2_{\b \Omega_{c_i,m}} \right) \right.\right. \\ & \left.\left. 
+ w_{\text{co}} \left\Vert \ \mrr_{\text{co}}(\check{\b{\mathcal{X}}},{\b{\mathrm{f}}}_m) \right\Vert^2
\right) 
+ \sum_{(i,j)\in\mathbb{I}}\sum^{N_l}_{l=1} \left\Vert \mrr_{\nu}(\check{\b{\mathcal{X}}},\b \nu^l_{ij})\right\Vert^2_{\b\Omega_{\nu_{ij},l}} \right. \\ & \left. +  \sum_{(i,j)\in\mathbb{I}} \left\Vert \mrr_\mathit{I}(\check{\b{\mathcal{X}}},\mathit{I}_{ij}) \right\Vert^2_{\b \Omega_{\mathit{I}_{ij}}} + \left\Vert   \mrr_\mathrm{\text{prior}}-\mrH_\mathrm{\text{prior}}\check{\b{\mathcal{X}}} \right\Vert^2  \right\},
\end{aligned}
\end{equation}
where~$\mrr_{c}(\cdot,\cdot)$, $\mrr_{\mathit{I}}(\cdot,\cdot)$, and~$\mrr_\mathrm{prior}$ are the visual, IMU, and prior factors of the VIKO module, respectively, with their corresponding covariance matrices denoted by~$\b\Omega$.~$\rho$ denotes the Huber norm as defined in~\cite{qin2018tro}.~$\mrH_\mathrm{\text{prior}}$ denotes the Jacobian matrix of the VIKO prior factor with respect to the sliding window states.~$\b{\nu}^l_{ij}$ and~$I_{ij}$ represent the foot velocity measurement of the~$l$-th foot and IMU measurement between the~$i$-th and~$j$-th~window, respectively.~${\b{\mathrm{f}}}_m$ and $N_c$ are the visual feature tracked from the stereo images and the number of visual features, respectively.~$\mrr_{\text{co}}(\cdot,\cdot)$, and~$\mrr_{\nu}(\cdot,\cdot)$ are the depth consistency and preintegrated foot velocity factors, respectively. 
$w_{\text{co}}$ is the depth consistency factor weight with value chosen according to the best result of each datasets.

We optimize both~\eqref{eq:ESIKF_cost_function} and~\eqref{eq:sliding_window_cost_function} alternately, as shown in~\figref{fig:factor_graph}, depending on the availability of camera or LiDAR measurements. Notice that the covariance of the states are retained in both optimizations, hence it is still a tightly-coupled framework. 

\subsubsection{LiDAR Factor}
\label{subsec:liko}
The LIKO module processes the incoming scan and matches it to the collected map using a voxel map~(\figref{fig:system_and_FactorGraph}(a)). Because each point in the scan is not recorded simultaneously and has its measurement time $t_p$, the points are deskewed to the current time $t_{k+1}$. The recorded propagated states from~\secref{subsec:imu_prop} are used to obtain the robot's rotation~${\b \mrR_{t_p}^w}$ and position~${\mrp}_{t_p}^w$ at~$t_p$. 
Subsequently, we can acquire each point's location in the current scan relative to the world frame, ${\b{\mathcal{P}}}^w_{n,t_{k+1}}$, by using known extrinsic between the IMU and LiDAR frames, ${\mrR}^{L}_{b}$ and ${\mrp}^{L}_{b}$, as follows:
\begin{equation}
\begin{aligned}
{\b{\mathcal{P}}}^w_{n,t_{k+1}}= \ & \hat{\mrR}^w_{k+1} \left({({\hat{\b \mrR}^w_{i})}^{\mathsf{T}}}{\b \mrR_{t_p}^w}({\mrR}^{L}_{b}{\b{\mathcal{P}}}^L_{t_p}+{\mrp}^L_{b}) \right. \\
& \left.+({{\hat{\b \mrR}^w_{k+1}}})^{\sf T}({\mrp}^w_{t_p}-\mrp^w_{t_\text{IMU}})\right)+\hat{\mrp}^w_{k+1},
\end{aligned}
\label{eq:deskew_point}
\end{equation}
where $t_{\text{IMU}}$ represents the nearest IMU measurement time relative to the latest LiDAR scan time. To compensate for the estimated pose, we define the residual as the distance from~$\b{\mathcal{P}}^w_{n,t_{k+1}}$ to its corresponding plane. The plane coefficient~$\mathbf{n}_{\mathcal{P}}$ is obtained using principal component analysis~\cite{pearson1901phisop} on the nearest surrounding points in the simultaneously built map. Finally, we define the residual of the LiDAR factor in~\eqref{eq:ESIKF_cost_function} as follows:
\begin{equation}
\mrr_\mathcal{P}(\check{\mathrm{\bf{x}}}_{k+1},\b{\mathcal{P}}^w_{n,t_{k+1}})={\normal_{\mathcal{P}}}^{\sf T}\left[\begin{array}{c} {\b{\mathcal{P}}}^w_{t_{k+1}} \\ 1 \end{array}\right].
\end{equation}

\begin{figure}[t!]
	\vspace{0.2cm}
	\captionsetup{font=footnotesize}
	\centering
	\def\svgwidth{0.45\textwidth}
	\graphicspath{{pics/040125_VoxelMap/}} 
	\input{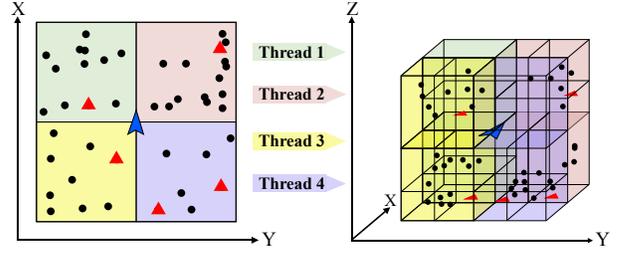}
	\caption{Illustration of our point cloud division for parallel voxel map update. Each color indicates a group of points assigned to a specific thread for insertion into the voxel map. The blue arrow and red triangle represent the current robot pose and newly observed points, respectively. Multiple parallel threads are utilized to insert the new points into the voxel map according to their spatial division.}
	\label{fig:pcl_division}
	\vsfig
\end{figure}
Additionally, we improved the point cloud insertion and voxel map management processes to minimize the optimization time by separating them into several parallel background threads. For point cloud insertion, rather than adding all the points from the current scan at once, we divide the point cloud into several segments and insert them concurrently into the map, as shown in~\figref{fig:pcl_division}. Similarly, the voxel map update is handled in the background to minimize the computational load on the main thread.

\subsubsection{Foot-preintegration}
\label{subsec:foot_pre} 
As shown in~\figref{fig:system_and_FactorGraph}(b), the preintegrated foot factor is utilized in both modules as additional residual for pose estimation. We leverage the measurement model from our previous study~\cite{yeeun2022icra} to obtain the latest foot velocities and positions. Suppose that~$\hat{\b \omega}^\xi_{l,i}$ and~$\hat{\b \nu}^\xi_{l,i}$ denote the obtained foot angular and linear velocities, respectively, in the foot frame~$\xi$ along with their respective noise terms~$\mrn_{{\b \omega}_{l,i}}$ and~$\mrn_{{\b \nu}_{l,i}}$. Each foot pose can be propagated according to the $i$-th joint measurement as follows:
\vspace{-0.3mm}
\begin{equation}
\begin{split}
\tilde{{\b \Psi}}^{w}_{l,i+1} ={\b \Psi}^{w}_{l,i} \mathrm{Exp}(\hat{\b \omega}^\xi_{l,i}-\mrn_{{\b \omega}_{l,i}})\Delta t, \\ \tilde{\mrs}^{w}_{l,i+1}=\mrs^{w}_{l,i}+{\b \Psi}^{w}_{l,i}(\hat{\b \nu}^\xi_{l,i}-\mrn_{{\b \nu}_{l,i})}\Delta t.
\end{split}
\label{eq:leg_prop}
\end{equation}
\vspace{-0.3mm}
where $\tilde{{\b \Psi}}^{w}_{l,i+1}$ and $\tilde{\mrs}^{w}_{l,i+1}$ denote the $l$-th foot's rotation and position propagation results, respectively.

Next, we construct the terms $\Delta{\tilde{\b \Psi}}_{l,ij}$ and $\Delta\tilde{\mrs}_{l,ij}$ for preintegrated rotation and position, respectively, by propagating all measurements within the current sliding window's timestamps. Using~\eqref{eq:leg_prop}, Lie algebra, and the first-order approximation~\cite{yeeun2022icra}, we can approximate the preintegrated measurement along with its noise as follows:
\begin{equation}
\small
\vsequ
\begin{split}
\Delta{\tilde{\b \Psi}}_{l,ij} & \cong \prod_{k=i}^{j-1}\left[\mathrm{Exp}(\hat{\b \omega}^{\xi}_{l,k} \Delta t) \ \mathrm{Exp}(-\mrJ_r^k\left(\hat{\b \omega}^{\xi}_{l,k} \Delta t\right)\mrn_{{\b \omega}_{l,i}}\Delta t)\right] \\
& =\Delta{\hat{\b \Psi}_{l,ij}} \ \mathrm{Exp}\left(-\delta \b{\psi}_{l,ij}\right),
\end{split}
\end{equation}
\begin{equation}
\begin{split}
\small
\Delta \tilde{\b{s}}_{l,ij} & \cong \sum_{k=1}^{j-1}\left[ \Delta \hat{\b \Psi}_{l,ik}\left( \mrI-\left(\delta\b{\psi}_{ik}\right)^{\wedge}\right)\left(\hat{\b \nu}^\xi_{l,k}-\mrn_{{\b \nu}_{l,i}}\right)\Delta t\right]\\
& \cong \sum_{k=1}^{j-1} \left[\Delta \hat{\b \Psi}_{l,ik} \hat{\b \nu}^\xi_{l,k} \Delta t\right]\\
& \ \ - \sum_{k=1}^{j-1}\left[\Delta \hat{\b \Psi}_{l,ik}\mrn_{{\b \nu}_{l,i}}\Delta t-\Delta \hat{\b \Psi}_{l,ik}\left(\hat{\b \nu}^\xi_{l,k}\right)^{\wedge}\delta\b{\psi}_{ik}\Delta t\right] \\
& = \Delta \hat{\b{s}}_{l,ij}-\delta \b{s}_{l,ij}.
\end{split} 
\label{eq:preintegrated_measurement}
\end{equation}
We use $(\cdot)^{\wedge}$ to denote the linear hat operator that maps a vector to Lie algebra. The noise vectors of preintegrated measurements~$\delta \b{\psi}_{l,ij}$ and~$\delta \b{s}_{l,ij}$ are utilized to obtain the preintegrated measurement covariance, which is incorporated into the optimization~\cite{yeeun2022icra}. Finally, we can construct the preintegrated foot velocity residual $\mrr_{\nu}(\check{\b{\mathcal{X}}},\b \nu^l_{ij})$ consisting of preintegrated position residual $\mrr_{\nu_{p}}(\cdot,\cdot)$ and rotation residual $\mrr_{\nu_{R}}(\cdot,\cdot)$ defined as follows:
\vsequ
\begin{equation}
\begin{aligned}
\mrr_{\nu_{R}}\left(\check{\b{\mathcal{X}}},\b 
{\hat{\omega}}^\xi_{l,ij}\right) & = \mathrm{Log}\left(\Delta\hat{\b{\Psi}}_{l,ij}^{\mathsf{T}}\Delta{\tilde{\b{\Psi}}}_{l,ij}\right), \\
\mrr_{\nu_{p}}\left(\check{\b{\mathcal{X}}},\b {\hat{\nu}}^\xi_{l,ij}\right) & = \mrr_{\nu_p}\left(\check{\mathrm{\bf{x}}}_{k+1},\b \nu^l\right) = \Delta \tilde{\b{s}}_{l,ij}-\Delta \hat{\b{s}}_{l,ij}. 
\end{aligned}
\end{equation}
Additionally, we optimize only the current pose in LIKO since the uncertainty information of the previous pose is unavailable.
\begin{figure}[t!]
	\captionsetup{font=footnotesize}
	\centering
	\def\svgwidth{0.48\textwidth}
	\graphicspath{{pics/0410_Framework/}} 
	\input{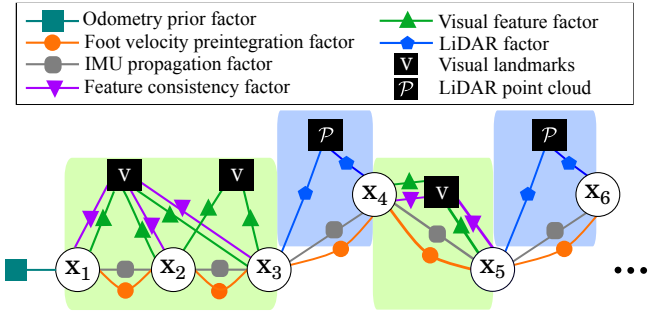}
	\caption{Optimization scheme of LVI-Q, which alternates between VIKO and LIKO module depending on the available measurements. The factors are combined by utilizing sliding window optimization and ESIKF estimation.}
	\label{fig:factor_graph}
	\vsfig
\end{figure}

For LIKO, the filter-based estimation requires the definition of partial derivative of $\mrr_{\nu_{p}}$ with respect to its error states. The derivative with respect to its error states is defined as follows:
\begin{equation}
\begin{aligned}
&\left. \partial\left(\mrr_{\nu_{p}}\left(\mrx\boxplus\delta\mrx,\b\nu^l\right) 
\boxminus \mrr_{\nu_{p}}\left(\mrx,\b{\nu}^l\right)\right) 
/ \partial \delta\mrx \right|_{\delta\mrx=0} \\
& \ \ \approx\left[\mrR\lfloor\b{s}_{l}\rfloor_{\times} \langle\mrr_{\nu_p}(\mrx,\b\nu^l)\rangle \ \langle\mrr_{\nu_p}(\mrx,\b\nu^l)\rangle \  \b{0} \ \dots \ \b{0} \ \mrR \right ],
\end{aligned}
\end{equation}
where $\lfloor\cdot\rfloor_{\times}$ and $\langle\cdot\rangle$ denote the skew symmetric matrix and normalization operators, respectively.

\subsubsection{Visual Factor}\label{subsec:viko}
\begin{figure}[t!]
	\vspace{0.25cm}
	\captionsetup{font=footnotesize}
	\centering
	\def\svgwidth{0.44\textwidth}
	\graphicspath{{pics/0918_New_FeatureAssc/}} 
	\input{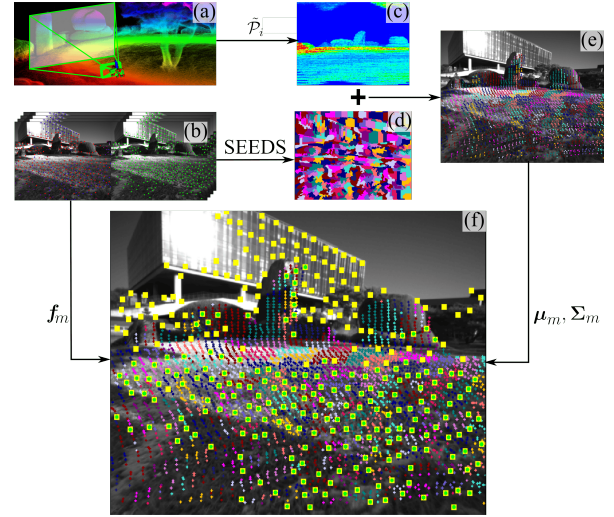}
	\caption{Visual feature consistency factor construction pipeline.~(a)~LiDAR point cloud utilized for depth consistency factor in the camera field of view.~(b)~Tracked visual feature points using FAST corner detector and KLT optical flow tracker on the image.~(c)~Projection of the LiDAR point cloud to the range image in the camera frame.~(d)~Result of the superpixel SEEDS~\cite{bergh2012eccv} algorithm to segment the image based on pixel information.~(e)~Grouped LiDAR points based on the superpixel segmentation shown by colored points.~(f)~Result of visual feature depth consistency according to its corresponding group. Yellow boxes with green dots indicate successful depth consistency factor generation (best viewed in color).}
	\label{fig:superpixel_point_cloud}
	\vsfig
\end{figure}

The well-established visual feature pipeline that manages a set of keyframes consisting of stereo images captured by a camera is employed. A FAST corner detector~\cite{roster2010tpami} and KLT optical flow tracker~(\figref{fig:superpixel_point_cloud}(b))~\cite{lucas1981IJCAI} are used to detect and track visual features across keyframes. We leverage the stereo reprojection error between the tracked points and their corresponding visual feature points. Assuming that we have tracked the visual feature in the left image plane~$^{c_l}{\b{\mathrm{f}}}_m$ and the corresponding right stereo pair~$^{c_r}{\b{\mathrm{f}}}_m$ with the pinhole projection model~$\b \pi_l(\cdot)$ and~$\b \pi_r(\cdot)$, the visual feature factor $\mrr_c(\check{\b{\mathcal{X}}},{\b{\mathrm{f}}}_m)=\left[\mrr_{c_l},\mrr_{c_r}\right]$ can be constructed as follows:
\begin{equation}
\begin{split}
 \mrr_{c_l}(\check{{\b{\mathcal{X}}}},{\b{\mathrm{f}}}_m) &= {^{c_l}\b{\mathrm{f}}}_m-{\b \pi}_l{(\b{\mathrm{f}}_m)}, \\
 \mrr_{c_r}(\check{\b{\mathcal{X}}},{\b{\mathrm{f}}}_m) &= {^{c_r}\b{\mathrm{f}}}_m-{\b \pi}_r(\b{\mathrm{f}}_m),\\
 {^{c_l}\b{\mathrm{f}}}_m &= \left(\hat{\mrR}^{w}_i {^{c_l}\hat{\mrR}^{b}_i}\right)^{\mathsf{T}} {\b{\mathrm{f}}}_m-\left({^{c_l}\hat{\mrR}}^{b}_i \right)^{\mathsf{T}} \mrp^w_i-^{c_l}\hat{\mrp}^b_i, \\
 {^{c_r}\b{\mathrm{f}}}_m &= \left(\hat{\mrR}^{w}_i {^{c_r}\hat{\mrR}}^{b}_i\right)^{\mathsf{T}} {\b{\mathrm{f}}}_m-\left({^{c_r}\hat{\mrR}}^{b}_i \right)^{\mathsf{T}} \mrp^w_i-^{c_r}\hat{\mrp}^b_i.
\end{split}
\end{equation}
\vspace{-4mm}

%

\subsubsection{Visual Consistency Factor}\label{subsubsec:visual_consistency} 
During deployment, we observed that depth estimation from stereo image matching frequently fails due to the bouncing motion of a legged robot. 
To address this, we introduce a more generalized approach to ensure the depth estimation consistency of ${\b{\mathrm{f}}}_m$ by fitting it to a normal distribution constructed from the surrounding points. Initially, we create superpixel~\cite{bergh2012eccv} clusters in the image~(\figref{fig:superpixel_point_cloud}(d)) based on the pixel information. We group the points within the image plane~$\tilde{\mathcal{P}}_i$ based on their respective spatial locations~(\figref{fig:superpixel_point_cloud}(e)). Utilizing the superpixel clusters containing visual features~(\figref{fig:superpixel_point_cloud}(f)), we calculate the corresponding 3D-NDT representation~\cite{magnusson2007jfr} by determining its mean and covariance. Each distribution is assumed to represent the characteristics of the underlying plane. Denoting $\left({\b{\mathrm{f}}_m}/{\lambda_m}-{\b \mu}_m\right)$ by $\b{\rm z}_m$, where $\b \mu_m$ is the mean of the NDT to which $\b{\mathrm{f}}_m$ belongs, the visual feature consistency factor is subsequently constructed by computing the negative \mbox{log-likelihood} of the point to distribution~\cite{magnusson2007jfr} as follows:
\vspace{-1mm}
\begin{equation}
\begin{aligned}
 \mrr_{\text{co}}(\check{\b{\mathcal{X}}},{\b{\mathrm{f}}}_m) = -a_1 \exp \left( -\frac{a_2}{2} \ {\b{\rm z}_m}^{\mathsf{T}} \b \Omega_m^{\inv}{\b{\rm z}_m}\right),
\end{aligned}
\end{equation}   
\begin{equation}
\begin{aligned}
 &a_1 =-\log(c_1+c_2)-a_3, \\
 &a_2 =-2\log\left( \left(-\log(c_1\exp(-1/2)+c_2)-a_3 \right)/a_1\right), \\
 &a_3=-\log(c_2),
 \end{aligned}
\end{equation}
where $\b \Omega_m$ is the covariance of the NDT to which $\b{\mathrm{f}}_m$ belongs; $c_1$ and $c_2$ denote the constants that bound the likelihood value.
\vspace{-0.8mm}

\newcommand\T{\rule{0pt}{2.6ex}}       
\newcommand\B{\rule[-1.2ex]{0pt}{0pt}} 

\section{Experimental Evaluation}
\label{sec:exp}
\subsection{Datasets}
\label{subcsec:datasets}
We extensively evaluated the proposed method on various datasets consisting of long trajectories and complex courses. First, we evaluate our system on the Fire Service College~(FSC)~\cite{camurri2020frontier} and CEAR~\cite{zhu2024ral} datasets, acquired using the \texttt{ANYmal} and \texttt{Cheetah} robots, respectively. In the FSC dataset, the \texttt{ANYmal} robot walks around a low-light construction site with wet concrete. For the CEAR dataset, we utilized the \texttt{mocap3{\_}blinking{\_}comb} (\texttt{mocap}) and \texttt{residential-area\_night\_comb} (\texttt{resident}) sequences where the \texttt{Cheetah} robot moves with a pronking gait in a room with blinking lights and dark environment, respectively. Both datasets present challenges in odometry when navigating dark and narrow environment. Additionally, the pronking gait in the~CEAR dataset introduces difficulty for contact-based kinematic factors because the robot rarely touches the ground during motion.  

\begin{figure}[t!]
	\captionsetup{font=footnotesize}
	\centering
	\def\svgwidth{0.5\textwidth}
	\graphicspath{{pics/0906_Quads/}} 
	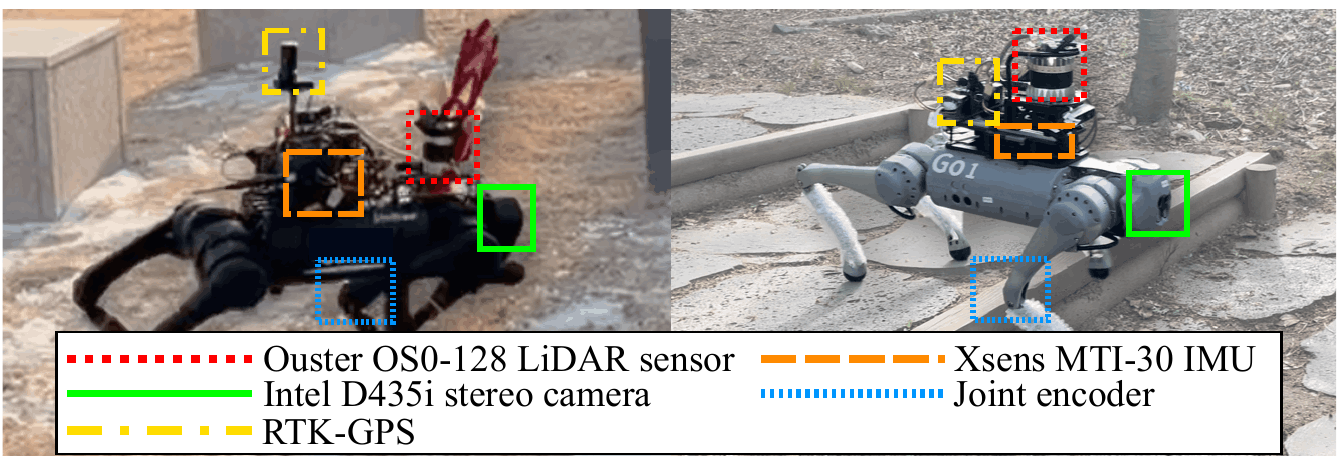
	\caption{Different quadruped robot platforms to acquire our own dataset. (L-R): Unitree A1 and Go1 robot platforms, respectively.}
	\label{fig:Quad}
	\vsfig
\end{figure}
In addition, we manually collected the K-Campus dataset, which consists of two sequences: \texttt{A1} acquired by the Unitree A1 robot~\cite{unitreeA1} and \texttt{Go1} by the Unitree Go1 robot~\cite{unitreeGo1}, as shown in~\figref{fig:Quad}. Both trajectories have a total length of over 800m and challenge the algorithms to estimate odometry through different environments in a single journey.
The RTK-GPS trajectory was used as a reference to quantitatively assess the resulting path accuracy. Additionally, an ablation study was performed by including only the foot-preintegration residual denoted by~\mbox{`+\textsl{Joint}'} and only the depth consistency factor denoted by~\mbox{`+\textsl{Depth consistency}'}.


For all datasets, except the \texttt{resident} sequence which does not provide the ground truth, we used the absolute trajectory and relative pose errors with reference to the available ground truth~\cite{yeeun2022icra}.  The resulting trajectory and ground truth are~SE(3)-aligned using the Umeyama's method for the first 500 poses~\cite{grupp2017git}. The EVO Python package~\cite{grupp2017git} was used to compute both errors. 


\subsection{Implementation Details}
\label{subsec:implementation_details}

\begin{table}[t!]
	\centering
	\captionsetup{font=footnotesize}
	\vspace{0.2cm}
	\caption{Absolute trajectory error~(ATE) and relative pose error~(RPE)~\cite{grupp2017git} on the quadruped robot. $\texttt{L}$, $\texttt{V}$, and $\texttt{K}$ indicate the modalities that are used for each approach, which represent LiDAR, visual, and kinematic information, respectively. All algorithms utilize the inertial information as motion prior. $\mathrm{Div.}$ indicates a divergence during execution.~(Units for ATE/RPE: m)}
	{\footnotesize \resizebox{\columnwidth}{!}{
			\begin{tabular}{l|ccc|cccc}
				\toprule
				\midrule
				{\multirow{4}{*}{\textbf{Method}}} & \multicolumn{3}{c|}{\multirow{2}{*}{Modality}} & \multicolumn{4}{c}{(ATE/RPE)  $\downarrow$}   \\ 
				& & & &  FSC Dataset & CEAR Dataset & \multicolumn{2}{c}{K-Campus Dataset} \\
				\cmidrule(lr){2-4} \cmidrule(lr){5-5} \cmidrule(lr){6-6} \cmidrule(lr){7-8} & \texttt{L} & \texttt{V} & \texttt{K} & \texttt{ANYmal} & \texttt{Cheetah} & \texttt{A1} & \texttt{Go1} \B\\ \midrule
				{VINS-Fusion~\cite{qin2019arxiv}} & & \checkmark & & \T 0.576/0.096 & 2.128/0.153 & 14.517/0.181 & 7.056/0.179 \B \\
				{Fast-LIO2~\cite{wei2022tro}} & \checkmark & & & \T 0.130/\textbf{0.095} & 0.175/0.137 & 1.189/0.136 & 3.164/0.142 \B \\
				{STEP~\cite{yeeun2022icra}} & & \checkmark & \checkmark & \T 0.329/\textbf{0.095} & 1.977/0.146 & 11.726/0.179 & 6.018/0.182 \B \\
				{FAST-LIVO~\cite{zheng2022iros}} & \checkmark & \checkmark & & \T 5.866/0.129 & 0.253/0.185 & 1.982/0.159 & 4.202/0.161 \B \\
				{R2LIVE}~\cite{lin2021ral} & \checkmark & \checkmark & & \T 0.317/0.108 & $\mathrm{Div.}$ & 4.866/0.144 & 5.321/0.146 \B \\
				{LVI-SAM}~\cite{shan2021icra} & \checkmark & \checkmark & & \T $\mathrm{Div.}$ & $\mathrm{Div.}$ & 8.100/0.267 & 4.761/0.258  \B \\
				{Cerberus}~\cite{yang2022icra} & & \checkmark & \checkmark & \T $\mathrm{Div.}$ & 2.002/0.151 & 21.963/0.136 & $\mathrm{Div.}$  \B \\
				\midrule
				\T
				+\textsl{Joint} & \checkmark & \checkmark &  \checkmark & 0.112/0.097 & 0.162/\textbf{0.134} & 0.867/0.138 & 1.278/0.144\B
				\\
				\T
				+\textsl{Depth consistency} & \checkmark &  \checkmark & & 0.112/0.105 & 0.204/0.138 & 0.981/0.141 & 1.565/0.144 \B
				\\
				\T
				\textsl{Ours} & \checkmark & \checkmark &  \checkmark & \textbf{0.108}/\textbf{0.095} & \textbf{0.159}/\textbf{0.134} & \textbf{0.756}/\textbf{0.135} & \textbf{1.173}/\textbf{0.137} \B
				\\
				\midrule
				\bottomrule
		\end{tabular}}
	}
	\label{tab:all_quads}
	\vstab
\end{table}
The proposed framework was built on the open-source R2LIVE~\cite{lin2021ral} and modified for compatibility with our platform. The deep reinforcement learning-based locomotion control algorithm,~\mbox{DreamWaQ~\cite{nahrendra2023icra}}, was employed as the controller. The testing platform consisted of a computer running Ubuntu Linux with an Intel(R) i7-12700k CPU. For the VIKO module, we set~$w_{\text{co}}$ to 
0.05.

\begin{figure*}[t!]
	\vspace{0.2cm}
	\captionsetup{font=footnotesize}
	\centering
	\def\svgwidth{0.93\textwidth}
	\graphicspath{{pics/1112_new_plot_over_map/}} 
	\input{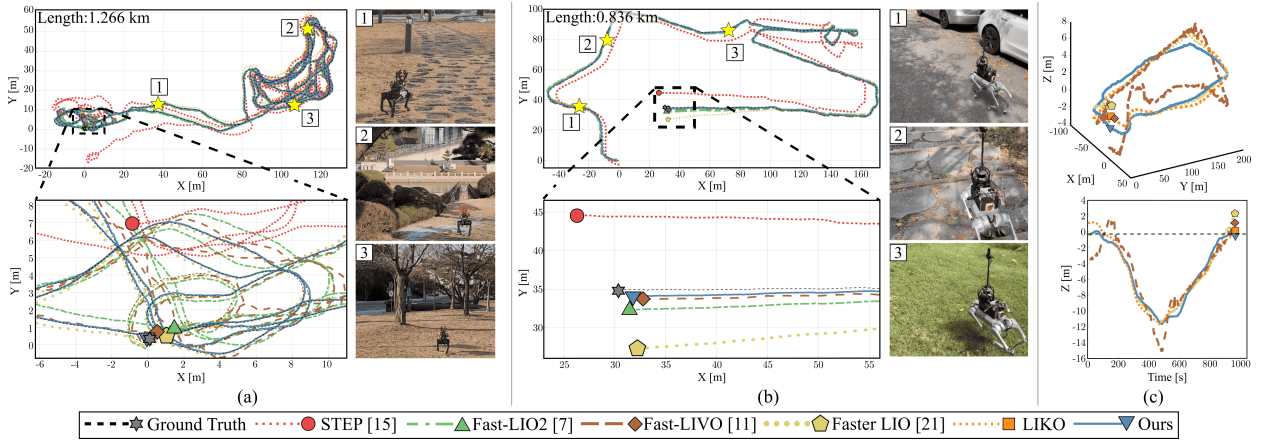}
	\caption{Qualitative comparison of (a)~\texttt{A1}, (b)~\texttt{Go1}, and (c)~\texttt{resident} sequences, respectively, showing that our method is better than other LIO~\cite{wei2022tro,bai2022ral}, VIKO~\cite{yeeun2022icra}, and LVIO-based~\cite{zheng2022iros} approaches. The closer the end point of ground truth and that of each algorithm, the better the performance. The symbol on each trajectory denotes the endpoint. All trajectories start from (0, 0, 0) and SE(3) aligned by the Umeyama's method for the first 500 poses~\cite{grupp2017git}.}
	\label{fig:Plot_over_map}
	\vsfig
\end{figure*}

\subsection{Comparison With State-of-the-Art Approaches}
\label{sec:result_comparison}
We conducted comparative analyses to compare our system with other open-source methods that utilize sensor setups similar to ours. VINS-Fusion\footnote{\href{https://github.com/HKUST-Aerial-Robotics/VINS-Fusion}{https://github.com/HKUST-Aerial-Robotics/VINS-Fusion}}~\cite{qin2019arxiv} is a renowned open-source odometry framework that utilizes a camera and IMU, frequently used in visual navigation. STEP~\cite{yeeun2022icra} and Cerberus\footnote{\href{https://github.com/ShuoYangRobotics/Cerberus}{https://github.com/ShuoYangRobotics/Cerberus}}~\cite{yang2022icra} utilize kinematic factors similar to ours but built on the visual-inertial odometry framework. Fast-LIO2\footnote{\href{https://github.com/hku-mars/Fast_LIO}{https://github.com/hku-mars/Fast\_LIO}}~\cite{wei2022tro} and Faster-LIO\footnote{\href{https://github.com/gaoxiang12/faster-lio}{https://github.com/gaoxiang12/faster-lio}}~\cite{bai2022ral} are the most popular odometry frameworks that use LiDAR and IMU sensors. R2LIVE\footnote{\href{https://github.com/hku-mars/r2live}{https://github.com/hku-mars/r2live}}~\cite{lin2021ral}, \mbox{LVI-SAM}\footnote{\href{https://github.com/TixiaoShan/LVI-SAM}{https://github.com/TixiaoShan/LVI-SAM}}~\cite{shan2021icra}, and~\mbox{Fast-LIVO}\footnote{\href{https://github.com/hku-mars/FAST-LIVO}{https://github.com/hku-mars/FAST-LIVO}}~\cite{zheng2022iros} are odometry frameworks that combine multiple modalities in a \mbox{tightly-coupled} manner comparable to ours. If a loop closure module was available in the algorithm, it was turned off for fair comparisons.

As shown in \tabref{tab:all_quads}, \figref{fig:Plot_over_map}(a), and \figref{fig:Plot_over_map}(b), \mbox{LVI-Q} achieves the lowest error compared with other algorithms. This superior performance is thanks to our tightly-coupled alternating estimation approach, which minimizes discrepancies between camera and \mbox{LiDAR} measurements. Our approach is able to accurately estimate the robot pose with more frequent updates (via factor-graph or filter-based optimization) and by leveraging different modalities that complement each other, leading to improved accuracy. \mbox{R2LIVE}, \mbox{Fast-LIVO}, and \mbox{LVI-SAM} experienced significant pose errors in large-scale environments due to drift from misaligned feature points during high-speed motion. Fast-LIO2 performed well but exhibited larger pose errors in large-scale scenes due to the continuous accumulation of drift. \mbox{VINS-Fusion} suffered severe drift caused by illumination changes or extrinsic errors, resulting in tracking failure or incorrect depth estimation. Despite incorporating kinematic data, Cerberus and STEP also showed significant performance degradation in large-scale environments because of long-term errors in visual depth estimation. These results confirm the robustness and generalization capabilities of~\mbox{LVI-Q}~across different platforms.

\begin{figure}[t!]
	\captionsetup{font=footnotesize}
	\centering
	\def\svgwidth{0.45\textwidth}
	\graphicspath{{pics/1213_dataset_map/}} 
	\input{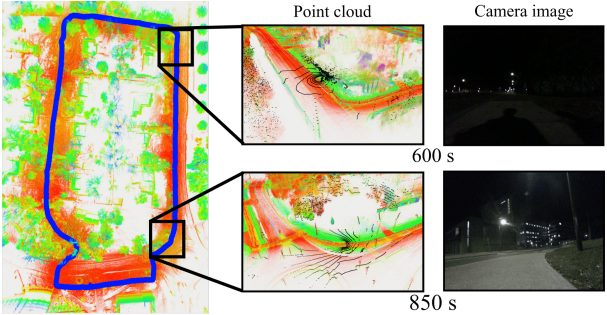}
	\caption{Point cloud map of \texttt{resident} sequence from LVI-Q system. VIO fails and LIO algorithms experience drift during estimation due to lack of light and geometrical features, especially at 600 and 850 sec in the sequence. Despite the challenges, LVI-Q successfully returns to the original position without any significant drift.}
	\label{fig:resident}
	\vsfig
\end{figure}

\begin{figure}[t!]
	\captionsetup{font=footnotesize}
	\centering
	\def\svgwidth{0.45\textwidth}
	\graphicspath{{pics/1219_new_VIKO/}} 
	\input{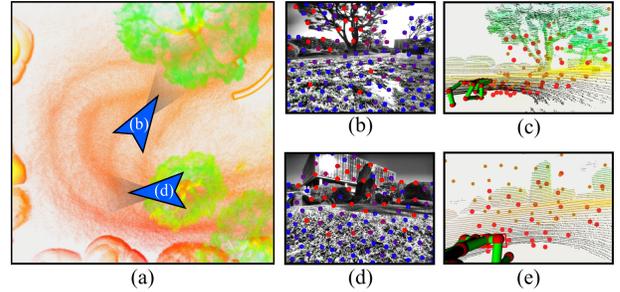}
	\caption{Tracked visual features visualized with its corresponding point cloud on the map in our K-Campus dataset. The blue arrows in (a) represent the robot's locations. The visual features are tracked from the images (b) and (d) containing bushes and trees. The red, brown, and blue dots represent tracked features optimized with depth consistency, tracked features optimized without depth consistency, and new visual features, respectively. The visual feature depth triangulation often fails in these scenes due to unstable tracking on ambiguous bush and tree corner areas. With our depth consistency factor, the depth of visual features are kept within its corresponding point cloud as shown in (c) and (e).}    
	
	\label{fig:illumination}
	\vsfig
\end{figure}

\subsection{Performance in Challenging Areas}
We tested our algorithm in challenging areas to ascertain our solution in preventing divergence, especially on the lack of visual features problem. \figref{fig:Plot_over_map}(c) shows the results of Faster-LIO~\cite{bai2022ral}, Fast-LIVO~\cite{zheng2022iros}, LIKO module, and \mbox{LVI-Q} on the \texttt{resident} sequence. Other VIO systems fail due to lack of tracked features in the dark environment, as shown in~\figref{fig:resident}. The Faster-LIO algorithm experienced drift because of little plane geometry in the outdoor environment, especially at the 600 and 850 sec of the sequence. \mbox{Fast-LIVO} shows unstable estimation results due to dark image patches that disturb the optimization process.~The LIKO module produces better estimation with less drift due to additional kinematic constraint.~\mbox{LVI-Q} is the only algorithm that returns to the same location without drift, especially on the \textit{z}-axis, as shown in~\figref{fig:Plot_over_map}(c). This result confirms our algorithm's capability that utilizes kinematic constraints to prevent incorrect pose estimation in a challenging environment. \\ \indent We also present an example of our optimized visual features in our K-Campus dataset with locations and viewpoints shown in~\figref{fig:illumination}(a). The visual features from images of bushes and trees (Figs. 9(b) and (d)) often have inaccurate depth triangulation due to unstable tracking. The original visual feature estimation without the depth consistency factor (denoted by brown dots) are scattered and is not correctly aligned with the point cloud. With the addition of our depth consistency factor, the resulting optimized visual features are aligned with the point cloud, as seen in~Figs. 9(c) and (e). This alignment leads to more accurate estimation results, as seen in~\tabref{tab:all_quads} with `+\textsl{Depth consistency}', and supports our contribution of improved depth consistency.

\subsection{Processing Speed}
\label{sec:processing_speed}
The image data were acquired at 15~Hz, whereas the LiDAR data were acquired at 10~Hz. The LIKO module constructs the LiDAR-kinematic factors in 10~ms, while the VIKO module calculates the visual-kinematic factors in~45 ms. After  evaluating the combined pipeline, our algorithm generates odometry at 11-13 Hz, ensuring real-time performance.

\section{Conclusion}
\label{sec:conclusion}

In this letter, we proposed the \textit{LVI-Q} odometry system, designed to efficiently integrate LiDAR, visual, inertial, and kinematic measurements into a unified framework. LVI-Q utilizes an alternating estimation process consisting of LIKO and VIKO modules that optimize the states based on available measurements. 
The LIKO module leverages a preintegration technique for foot velocity measurements obtained from joint sensors, significantly improving height estimation. Meanwhile, the VIKO module utilizes the depth consistency factor derived from LiDAR point cloud data to refine visual feature depth values.
Extensive experiments across various platforms and large-scale environments demonstrated the robustness of our proposed \mbox{LVI-Q}.

Future work could enhance the algorithm by using deep learning methods to track foot poses from sequential joint measurements as well as generating depth maps of visual features to improve camera and LiDAR alignment.



\bibliographystyle{URL-IEEEtrans}

\bibliography{URL-bib}

\end{document}